\documentclass{article}
\usepackage{spconf,amsmath,graphicx}
\usepackage{amsthm}
\usepackage{amsfonts,bm}
\usepackage{float}
\usepackage{multirow}
\newcommand{\tabincell}[2]{\begin{tabular}{@{}#1@{}}#2\end{tabular}}

\title{Self-Attention Aligner: A Latency-Control End-to-End Model for ASR Using Self-Attention Network and Chunk-Hopping}
\name{Linhao Dong$^{1,2}$, Feng Wang$^1$, Bo Xu$^1$
}
\address{
$^1$Institute of Automation, Chinese Academy of Sciences, China\\
$^2$University of Chinese Academy of Sciences, China\\
\small \tt \{donglinhao2015, feng.wang, xubo\}@ia.ac.cn}
\begin{document}
\ninept
\maketitle
\begin{abstract}
Self-attention network, an attention-based feedforward neural network, has recently shown the potential to replace recurrent neural networks (RNNs) in a variety of NLP tasks. However, it is not clear if the self-attention network could be a good alternative of RNNs in automatic speech recognition (ASR), which processes the longer speech sequences and may have online recognition requirements. In this paper, we present a RNN-free end-to-end model: self-attention aligner (SAA), which applies the self-attention networks to a simplified recurrent neural aligner (RNA) framework. We also propose a chunk-hopping mechanism, which enables the SAA model to encode on segmented frame chunks one after another to support online recognition. Experiments on two Mandarin ASR datasets show the replacement of RNNs by the self-attention networks yields a 8.4\%-10.2\% relative character error rate (CER) reduction. In addition, the chunk-hopping mechanism allows the SAA to have only a 2.5\% relative CER degradation with a 320ms latency. After jointly training with a self-attention network language model, our SAA model obtains further error rate reduction on multiple datasets. Especially, it achieves 24.12\% CER on the Mandarin ASR benchmark (HKUST), exceeding the best end-to-end model by over 2\% absolute CER.
\end{abstract}
\section{Introduction}
\label{sec:intro}
End-to-end models \cite{graves2012sequence, chan2016listen, jaitly2016online, raffel2017online, sak2017recurrent} have greatly simplified the automatic speech recognition (ASR) system by combining acoustic model, language model and an acoustic-to-text alignment mechanism (e.g. attention \cite{chan2016listen, jaitly2016online, raffel2017online}, CTC-like \cite{graves2012sequence, sak2017recurrent}) in a unified neural network. As a common component of these models, recurrent neural networks (RNNs) \cite{graves2013speech, sak2014long, chung2014empirical} have demonstrated their sequential modelling power in both of capturing acoustic dependencies (acoustic modelling) and recurrently emitting text units (language modelling). However, RNNs may generate ``confusing'' internal states (memory) after passing through noisy pieces (e.g. long silence or noise pieces in speech utterances). Besides, the sequential nature of RNNs leads to low parallelization and slow computation speed. These shortcomings may restrict the performance and efficiency of the RNN-based end-to-end models in ASR task.

Recently, an attention-based feedforward neural network, called self-attention network, has shown promising performance in a variety of NLP tasks including neural machine translation \cite{vaswani2017attention}, reading comprehension \cite{yu2018qanet}, etc. This network captures positional dependencies of a sequence by computing pairwise attention weights, which could be small so as to bypass the unrelated (e.g. noisy) positions, thus it may leverage the related context information more effectively. In addition, the self-attention network models in a totally feedforward manner, thus providing highly paralellizable computation. Those advantages make it a potential alternative of RNNs.

However, there are some challenges in the replacement of RNNs by the self-attention networks for end-to-end modelling in ASR. Firstly, speech sequences often contain hundreds of, even over one thousand frames, it is not clear how the self-attention network could better encode in such a long range. Secondly, the self-attention network decodes output units in an auto-regressive manner, it is unclear if it could be effectively combined with the CTC-like alignments or an extra language model (LM). Thirdly, the self-attention network computes by relating all pairwise positions in a sequence, which means the entire utterance needs to be obtained at first, thus bringing difficulties for online recognition. To explore above challenges, we introduce the self-attention networks to a simplified recurrent neural aligner (RNA) framework \cite{dong2018extending}. Our contributions are as follows:

\begin{itemize}
\item We construct an encoder that relies only on self-attention and shallow convolutional networks. Pooling layers in between the self-attention networks plus the front-end strided convolutions, offering effective temporal down-sampling for speech utterances. A 5.5\% relative character error rate (CER) reduction on the HKUST dataset demonstrates the superiority of our encoder than a strong RNN-based encoder.
\item We present a self-attention decoder, which emits output units in an auto-regressive manner. It works well with the CTC-like alignment and provides a 2.4\% relative CER reduction than a RNN-based decoder.
\item We combine the proposed encoder and decoder for end-to-end training, and term the integrated model as self-attention aligner (SAA). We find the SAA model performs competitive on two Mandarin ASR datasets. Moreover, after jointly training with a self-attention network LM, it obtains further performance benefits.
\item We propose a chunk-hopping mechanism, which enables the SAA to support online speech recognition. Results show the chunk-hopping allows the SAA model to have only a 2.5\% relative CER degradation with a 320ms latency, increasing the diversity of application scenarios for the SAA model.
\end{itemize}

\section{Relations to prior work}
Self-attention network has been applied to ASR community in several prior works \cite{povey2018time, dong2018speech, zhou2018comparison, sperber2018self}. In \cite{povey2018time}, Povey et al. proposed a time-restricted self-attention layer, which improves the performance of the LF-MMI model when combining with the TDNN or TDNN-LSTM structure. In \cite{dong2018speech, zhou2018comparison}, the self-attention network is utilized in the transformer framework, which entirely relies on the attention mechanism and transcribes speech utterances in a sequence-to-sequence manner. In \cite{sperber2018self}, Sperber et al. applied the self-attention network to the encoder of the LAS model \cite{chan2016listen} and proposed several improvements for effective acoustic modelling of self-attention.

In this paper, we aim to explore the combination of the self-attention networks with the CTC-like alignment mechanism, differing from the HMM alignment in \cite{povey2018time} or the attention alignment in \cite{dong2018speech, zhou2018comparison, sperber2018self}. The CTC-like alignment mechanism provides the potential of online recognition, which is what above attention-based models lack. Besides, we utilize multiple vanilla self-attention networks in \cite{vaswani2017attention}, contrasting to a single time-restricted self-attention layer in \cite{povey2018time}, which is placed towards the end of the TDNN or TDNN-LSTM and provides latency-control by attending to limited future high-level context. In contrast, our model attends to segmented input frame chunks one after another, thus controlling the latency more directly without considering the setting of used neural networks.

\section{Self-attention Network}
Self-attention is an attention mechanism that computes the representation of a single sequence by relating different positions in it.
%Recently, there have appeared different types of self-attention mechanisms including the scaled dot-product self-attention \cite{vaswani2017attention}, the multi-dimensional self-attention \cite{shen2017disan}, etc.
In this work, we employ the scaled dot-product self-attention in the transformer \cite{vaswani2017attention}, and leverage its encoder block as the self-attention network (SAN), which contains two sub-networks: multi head self-attention and position-wise feed-forward network. In addition, the layer normalization, dropout and residual connection in the SAN are also introduced for effective training.

Let $X \in \mathbb{R} ^{T \times d}$ be an input sequence, where T is the sequence length and $d$ is the hidden size of the SAN. Let $X_1$, $Y_1$ be the input and output of the first sub-network: multi-head self-attention network, $X_2$, $Y_2$ be the input and output of the position-wise feed-forword network. The computation of the SAN is formulated as follows:
\begin{equation}
X_1=\text{LayerNorm}(X)
\end{equation}
%\begin{equation}
%Q=X_1W^Q, K=X_1W^K, V=X_1W^V
%\end{equation}
\begin{equation}
\begin{split}
\text{head}_i = \text{softmax}(\frac{Q_iK_i^T}{\sqrt{d/h}}+\text{bias}) V_i \quad\text{(i = 1, 2, ..., h)} \\
\text{where~~} Q_i=X_1W^Q_i, K_i=X_1W^K_i, V_i=X_1W^V_i
%\text{where~~} Q_i, K_i, V_i = \text{get\_ith\_state}(\text{split}(Q, K, V, h))
\end{split}
\end{equation}
\begin{equation}
Y_1 = \text{Concat}(\text{head}_1,\text{head}_2,\ldots,\text{head}_h)W^O
\end{equation}
\begin{equation}
X_2 = \text{LayerNorm}(\text{Dropout}(Y_1)+X_1)
\end{equation}
\begin{equation}
Y_2 = \text{max}(0, X_2W_1+b_1)W_2+b_2
\end{equation}
\begin{equation}
SAN(X) = \text{Dropout}(Y_2)+X_2
\end{equation}
Where, h is the number of heads in the multi-head self-attention network, which jointly attends to the information from different subspaces mapped by $W^Q_i$, $W^K_i$, $W^V_i \in \mathbb{R} ^{d, d/h}$. Each head relates different positions by computing pairwise dot-product values, which are scaled and then added with the $bias \in \mathbb{R} ^{T \times T}$ for affecting the attention manner. Since the speech signal is consecutive, in this work, we encourage attention to closer positions by adding the proximity bias: $\text{-log(}1+a\text{)}$ to each position-pair with distance $a$. Equation (5) represents the computation of the position-wise feed-forward network, which consists of two linear transformations with a ReLU activation in between, the inner layer has dimensionality $d_{ff}$. Dropout appears in equation (4), (6) is the residual dropout. Another dropout, called the attention dropout, is applied to the softmax weights in equation (2) but not showing in the equation.

\section{Model Architecture}
\subsection{Self-attention Aligner}
Self-attention aligner (SAA) is a RNN-free end-to-end model that contains three parts: an encoder, a decoder and a CTC-like alignment mechanism. The encoder transforms speech features, which can also be depicted as 2-dimensional spectrograms, to high-level acoustic representations. Then, with the joint action of the CTC-like alignment mechanism, the decoder learns to predict output units (e.g. characters, word pieces) by leveraging the encoded acoustic information and previous decoded outputs. The detailed model architecture is illustrated in figure 1.

\begin{figure}[h]
  \centering
  \vspace{-2.0mm}
  \includegraphics[width=\linewidth]{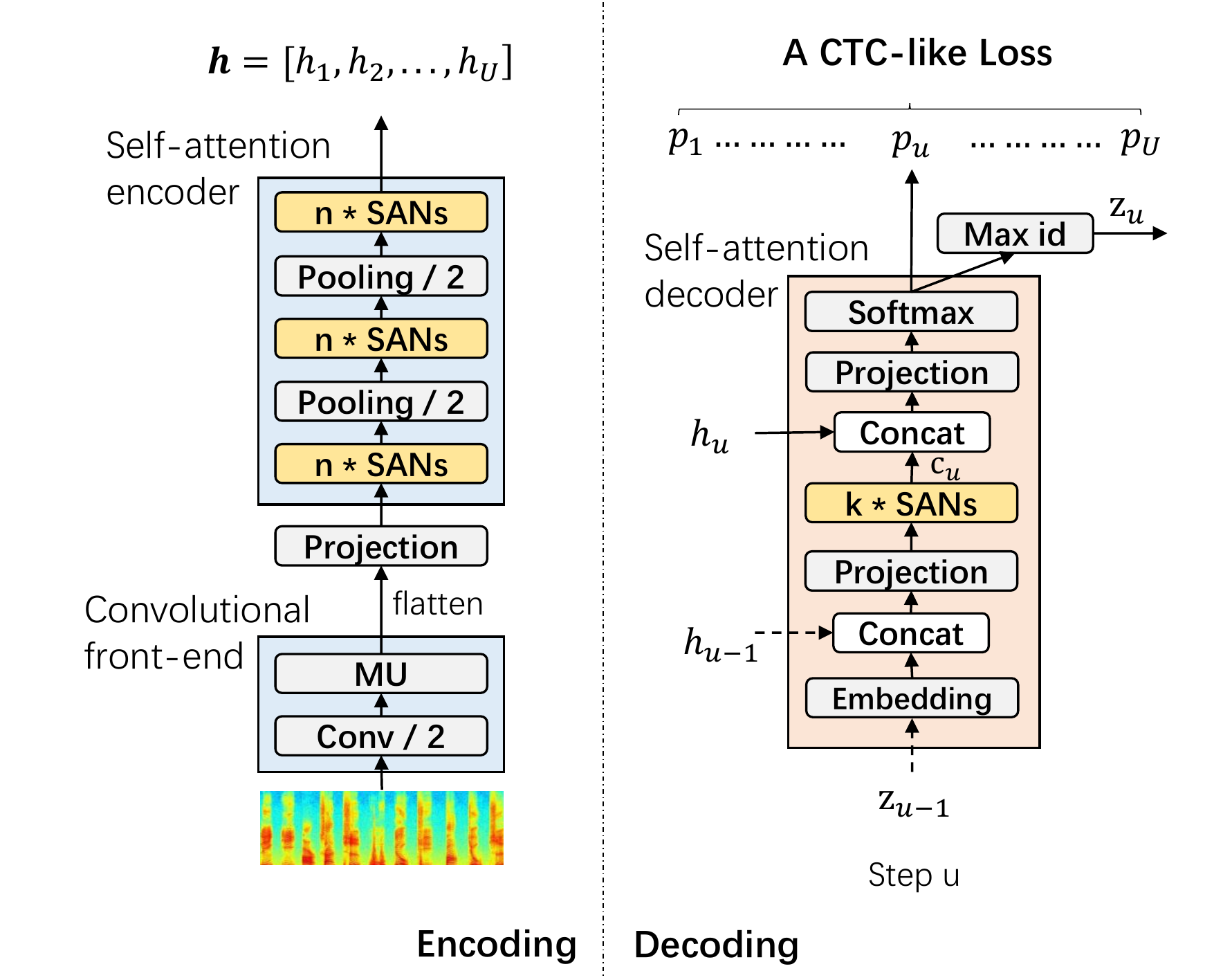}
  \vspace{-4.0mm}
  \caption{The model architecture of self-attention aligner.}
  \label{fig:SelfAttentionAligner}
  %\vspace{-5.0mm}
  \vspace{-2.0mm}
\end{figure}

The encoder, as shown in the left half of figure 1, transforms speech sequences only by self-attention and shallow convolutional networks. The convolutional front-end, employing the same structure in \cite{dong2018extending}, utilizes a strided convolutional layer to offer translational invariance while halving sequence length, and a multiplicative unit (MU) to further capture distinguishable acoustic details. Then, its 2-dimensional outputs are flatten and projected to representations with hidden size $d$ to as the input of self-attention encoder, which consists of stacked self-attention networks (SANs). Since the proximity bias in the SANs has provided the relative position information, we abandon the sinusoidal position encoding in \cite{vaswani2017attention}. Besides, between the stacked SANs, we place temporal pooling layers to conduct down-sampling, the motivation behind is as two-folds: (1) It encourages effective encoding in different temporal resolution. (2) It further shortens the length of acoustic representations, thus promoting faster alignments in the decoding. After the entire encoding, acoustic representations $\bold{h} = (h_1, \ldots, h_u, \ldots, h_U)$ are obtained.

The decoder, illustrated in the right half of figure 1, is also computed using stacked SANs. Differing from the application in the encoder, the SANs in the decoder are computed in an auto-regressive manner, which restricts each position to attend to positions up to and including that position. Thus, at step $u$, we cache the computed self-attention states $K_i$, $V_i \in \mathbb{R} ^{u-1, d/h}$ of all heads for the dependency modelling of later positions. Additionally, in order to make full use of acoustic information in the decoding, we concatenate $h_{u-1}$ with the embedding of previous predicted label $z_{u-1}$ to as the input of self-attention decoder. Besides that we also concatenate $h_u$ with the output of the SANs and project to the logits with size $L+1$, where L means the number of real output labels, and the extra one means the blank label, which is used for the acoustic-to-text alignment.

The alignment mechanism, aims to find an alignment $\bold{z} = (z_1, \ldots, z_u, \ldots, z_U)$ between the acoustic representations $\bold{h} = (h_1, \ldots, h_u, \ldots, h_U)$ and the target sequence $\bold{y} = (y_1, \ldots, y_n, \ldots,\\ y_N)$. Here, we utilize a simplified RNA alignment mechanism \cite{dong2018extending}, its conditional distribution $p(\bold{z}|\bold{h})$=$\prod_{u}p$($z_u|z_{u-1}, \bold{h}$), where $z_{u-1}$ is the label with the maximum probability at previous step. This mechanism simplifies the computation of the RNA decoder \cite{sak2017recurrent}, meanwhile keeping the computation consistency during training and inference. The loss function to be minimized is calculated by:
\begin{equation}
  \mathcal{L}=-\sum_{(\bold{h}, \bold{y})}\text{log }p(\bold{y}|\bold{h}) = -\sum_{(\bold{h}, \bold{y})}\text{log}(\sum_{\bold{z} \in {\beta}^{-1}(\bold{y})}p(\bold{z}|\bold{h}))
  \label{eq3}
\end{equation}
where $\beta$ is a ``CTC-like mapping function'', which maps the alignment $\bold{z}$ to the corresponding $\bold{y}$ by just removing the blank labels.

\subsection{Joint Training with a SAN-LM}
We find self-attention network language model (SAN-LM) obtains better perplexity than recurrent neural network language model (RNN-LM) at the character-level (the details are in section 5.2). In order to leverage more effective language information, we combine the SAA model with a pre-trained SAN-LM by a joint training mechanism similar to \cite{dong2018extending}.

\begin{figure}[h]
  \centering
  \vspace{-2.0mm}
  \includegraphics[width=\linewidth]{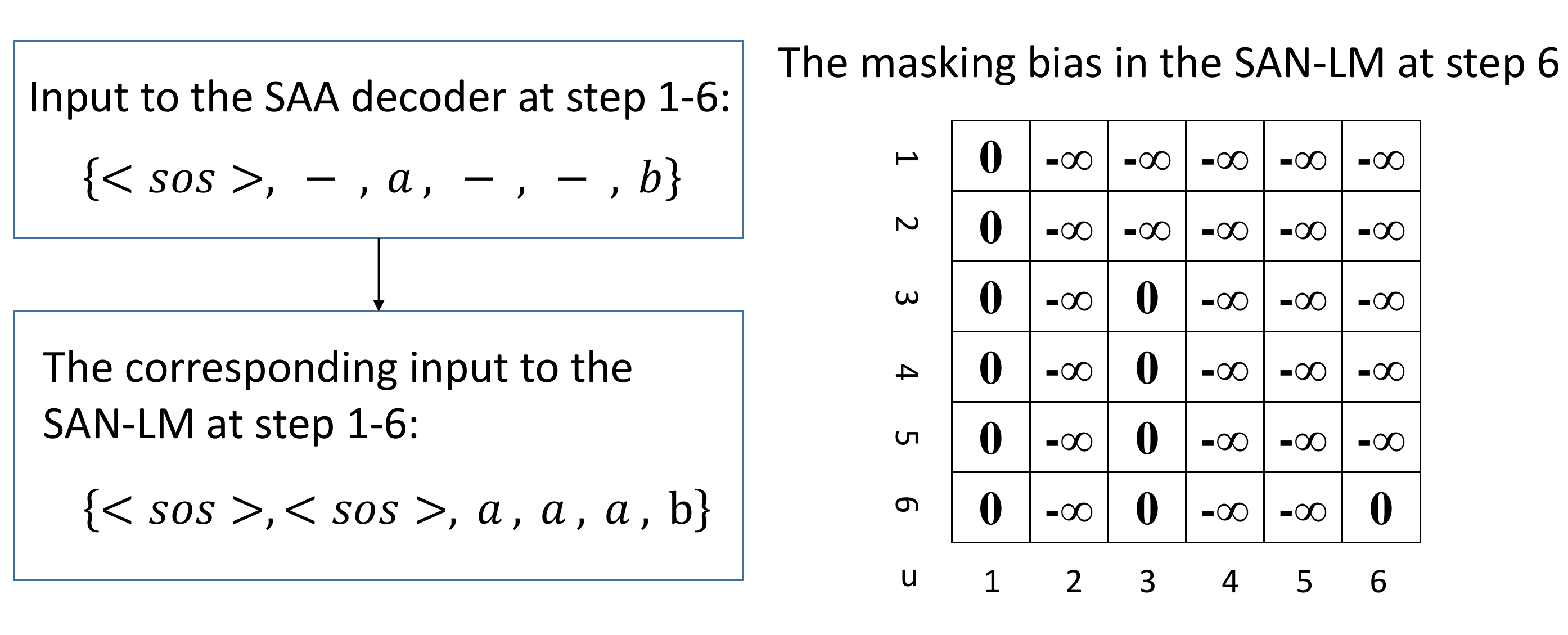}
  \vspace{-6.0mm}
  \caption{An example of the setting for the SAN-LM in joint training. ``$-$'' (in the left part) represents the blank, ``$-\infty$'' (in the right part) is used to mask out the attention between illegal position-pairs.}
  \label{fig:sanlm}
  \vspace{-1.0mm}
\end{figure}

At each step $u$, the predicted label $z_{u-1}$ is used as the input to the SAA decoder and the SAN-LM to calculate the corresponding SAA state and LM state, respectively. However, $z_{u-1}$ is likely to be the blank label, which is not seen in the training of the SAN-LM. Thus we let $z_{u-1}=z_{u-2}$ if $z_{u-1}$ is the blank label, and let $z_0=$\textless{sos}\textgreater \, which represents a special label of start for all sentences. In the calculation of the LM state, we introduce a masking bias to make the SAN-LM just attend to positions whose original $z_{u-1}$ is non-blank. We also abandon the proximity bias in the SAN-LM due to the changed positional distance between the separate LM training and the joint training. An example of above setting is illustrated in figure 2. After obtaining the LM state, we follow the same fusion structure as \cite{dong2018extending} to get the logits, and only the fusion structure is optimized during the joint training.

\subsection{Chunk-hopping Mechanisms}
The calculation of SANs needs to relate all pairwise positions in a sequence, which makes it necessary for the SAA model to start recognizing after entire utterance has been obtained. For this problem, we propose a chunk-hopping mechanism, which enables the SAA to support online recognition by encoding on segmented frame chunks sequentially (shown in the figure 3).

\begin{figure}[t]
  \centering
  \includegraphics[width=\linewidth]{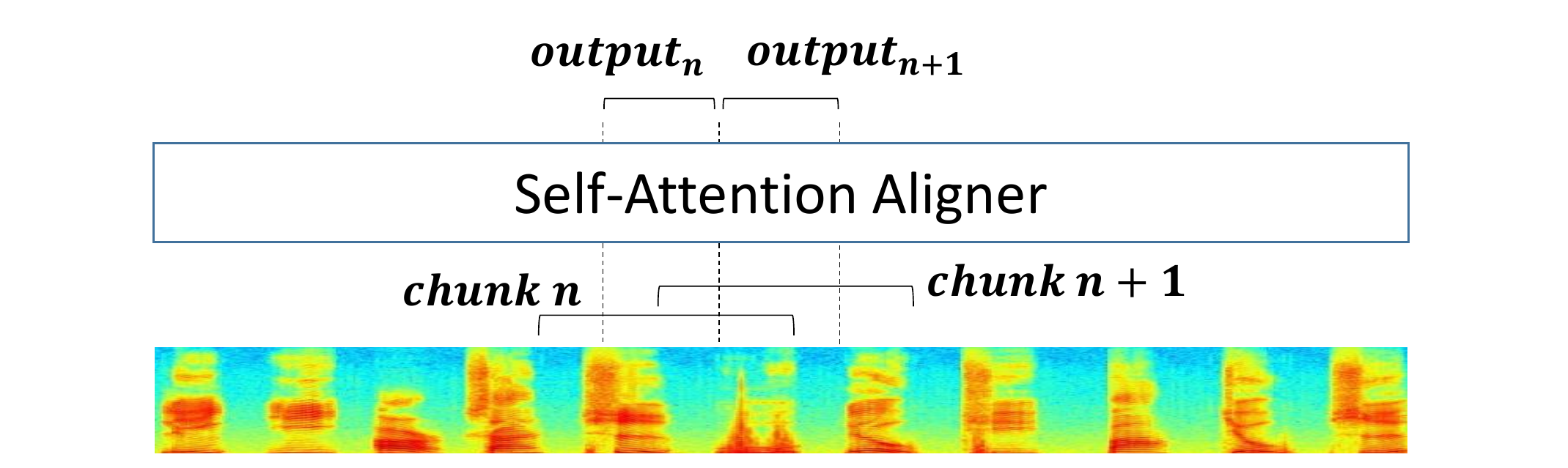}
  \vspace{-6.5mm}
  \caption{The illustration of chunk-hopping mechanism.}
  \label{fig:chunkhopping}
  \vspace{-5.5mm}
\end{figure}

We first segment entire utterance into several overlapped chunks, each of them contains three parts: one of which is the current part, whose output is used as the output of the chunk. Other two parts are the past, future part, which provide contexts for the calculation of the current part. After calculating one chunk, a hopping is generated for the recognition of the next chunk, and the hop size is equal to the size of the current part in each chunk. When calculating the beginning and end chunks, zeros are padded to make them work.

\section{Experiments}
\label{sec:experiments}

\subsection{Experimental Setups}
\label{ssec:experiments}
We experiment with two Mandarin Chinese conversational telephone speech recognition (MTS) datasets, including the Mandarin ASR benchmark (HKUST) \cite{liu2006hkust} and a larger-scale dataset (CasiaMTS).

The HKUST has 5413 utterances ($\sim$5 hours) for evaluation, we extract 6017 utterances ($\sim$5 hours) as our development set from the original training set with 197387 utterances ($\sim$173 hours) and use the left as our training set. Input features use 40-dimensional filterbanks extracted from a 25ms window and shifted every 10ms, extended with delta and delta-delta, then with the per-speaker and global normalization. Output units contain 3673 classes, including 3642 Chinese characters, 26 lowercase letters, 3 special character (noise etc.), the \textless{sos}\textgreater \ label and the blank label. In the convolutional front-end, the filter number is set to 64, and layer normalization \cite{ba2016layer} is applied after the convolutions. In the self-attention encoder and decoder, the hidden size $d=320$, the head number $h=4$, the residual dropout and the attention dropout is set to 0.1, the inner size $d_{ff}$ is set to 1280 except the experiments on the augmented data, is set to 2560. The confidence penalty regularizer in \cite{dong2018extending} is also introduced with the hyper-parameter $\lambda=0.2$. The SAN-LM used in our experiments contains 3 self-attention networks, with the same $d$, $h$, $d_{ff}$ as the SAA, but with larger dropout value 0.2. The RNN-based baseline uses the Extended-RNA model in \cite{dong2018extending}, which leverages 4-layers bidirectional LSTM (BLSTM) \cite{Schuster1997Bidirectional} as the encoder and 1-layer LSTM \cite{Hochreiter1997Long} as the decoder. The RNN-LM also follows the setting in \cite{dong2018extending}, utilizes 1-layer LSTM with 640 cells.

The CasiaMTS has four representative test sets which contain 1315, 967, 2280, 17793 utterances, respectively. The development set contains 20000 utterances and the train set has 1109696 utterances ($\sim$745 hours). Output units contain 4622 classes, including 4594 Chinese characters, 26 uppercase letter, the \textless{sos}\textgreater \ label and the blank label. We directly utilize the same SAA model as the HKUST dataset except the output layer becomes to 4622 units.

\subsection{Results}
\label{set:results}
We first explore the effects of replacing RNNs by the SANs. The corresponding results are shown in table 1, where n, k have the same meaning as in figure 1, specifically, n represents the number of the SANs at each temporal resolution of the encoder, k represents the number of the SANs in the decoder. We find replacing the LSTM-encoder, LSTM-decoder in the RNN-based baseline with our SAN-encoder, SAN-decoder yields a 5.5\%, 2.4\% character error rate (CER) reduction, respectively. We also find the SAA model performs better as the number of the SANs start increasing, but after increasing to a certain number of layers, improvements become limited or even decreased. In the later part, we use the best performing model with $n$=5, $k$=2 in table 1 as the default SAA model.

\begin{table}[!ht]
\centering
\vspace{-2.5mm}
\caption{The performance effects of replacing RNNs with SANs. (Unless otherwise state, results are on the HKUST dataset.) }
\vspace{1.5mm}
\begin{tabular}{c|c|c|c}
\hline
Model name & n & k & CER \\
\hline \hline
\tabincell{c}{Baseline \cite{dong2018extending}\\(LSTM-encoder + LSTM-decoder)} & - & - & 28.07 \\
\hline \hline
\multirow{4}{*}{SAN-encoder + LSTM-decoder} & 3 & - & 27.25 \\
& 4 & - & 26.86 \\
& 5 & - & 26.51 \\
& 6 & - & 26.41 \\
\hline \hline
\multirow{4}{*}{\tabincell{c}{SAA model \\ (SAN-encoder + SAN-decoder)}} & 5 & 1 & 26.24 \\
& 5 & 2 & \textbf{25.88} \\
& 5 & 3 & 26.01 \\
& 5 & 4 & 26.14 \\
\hline
\end{tabular}
\vspace{-0.5mm}
\end{table}

Besides the performance improvements, our SAA model also obtains speed improvements not only in the training stage but also in the inference stage (in table 2), showing the advantages of the replacement of RNNs by SANs.

\begin{table}[!ht]
\centering
\vspace{-2.5mm}
\caption{The speed comparison between the SAA model and the RNN-based baseline.}
\vspace{1.5mm}
\begin{tabular}{c|c|c}
\hline
Model name & steps/sec (training) & utts/sec (inference) \\
\hline \hline
Baseline \cite{dong2018extending} & 0.318 & 27.48 \\
SAA model & \textbf{0.415} & \textbf{31.47} \\
\hline
\end{tabular}
\vspace{-0.5mm}
\end{table}

Then, we compare the performance of jointly training with different language models for the SAA model (in table 3), and find the SAN-LM not only obtains lower perplexity but also provides more extra language information for the SAA model. The SAA combined with the SAN-LM is used for further comparison with other results.

\begin{table}[!ht]
\centering
\vspace{-2.5mm}
\caption{The comparison of incorporating different language models into the SAA model by the joint training.}
\vspace{1.5mm}
\begin{tabular}{c|c|c}
\hline
LM type & LM perplexity & CER\\
\hline \hline
No LM & - & 25.88 \\
\hline \hline
RNN-LM & 44.3 & 25.11 \\
SAN-LM & 43.1 & \textbf{24.92} \\
\hline
\end{tabular}
\vspace{-0.5mm}
\end{table}

Table 4 shows the investigation on the chunk-hopping mechanism, except the row 8, the number of frames in the past and future part keeps the same. We first compare different chunk sizes (row 2-4) under the same hop size 32. In line with the intuition, the better result is obtained under the wider chunk. Then, under the same chunk size, we compare the performance obtained by different hop sizes (row 4-7), and find the hop size 64 performs the best, which addresses the importance of suitable context information. Next, we widen the chunk size to 192 to further explore the effects brought by the changing of contexts, and find widening the past and future parts (row 9) at the same time performs better than only widening the past part (row 8). Even so, the chunk-hopping setting in row 8 achieves a 26.52\% CER, a 2.47\% degradation compared with the full-sequence result with a latency of 320ms.

\begin{table}[!ht]
\centering
\vspace{-0.5mm}
\caption{The results of equipping with different chunk-hopping setting for the SAA model (all models are trained from scratch).}
\vspace{1.5mm}
\begin{tabular}{c|c|c|c|c}
\hline
use & chunk size & hop size & future size & CER \\
\hline \hline
no & - & - & - & 25.88 \\
\hline \hline
yes & 32 frames & 32 frames & 0 frames & 31.99 \\
yes & 64 frames & 32 frames & 16 frames & 28.80 \\
yes & 128 frames & 32 frames & 48 frames & 27.15 \\
\hline \hline
yes & 128 frames & 64 frames & 32 frames & 27.09 \\
yes & 128 frames & 96 frames & 16 frames & 27.57 \\
yes & 128 frames & 128 frames & 0 frames & 28.53 \\
\hline \hline
yes & 192 frames & 64 frames & 32 frames & 26.52 \\
yes & 192 frames & 64 frames & 64 frames & 26.28 \\
\hline
\end{tabular}
\vspace{-0.5mm}
\end{table}

Table 5 lists the comparison between the SAA model and other published models \cite{watanabe2018espnet, dong2018extending, zhou2018comparison, povey2016purely} on the HKUST dataset. To our best knowledge, the results of all comparison models in table 5 are the latest. For a fair comparison, we also augment the training data by linearly scaling the audio lengths by factors of 0.9 and 1.1 (speed perturb), which brings a 0.8 absolute CER reduction. Finally, our SAA model obtains a 24.12\% CER, which exceeds the best end-to-end results from transformer by over 2\% absolute CER, but still has a little performance gap from the LF-MMI model, which uses the left-to-right alignment of the HMM rather than an end-to-end alignment.

\begin{table}[!ht]
\centering
\vspace{-2.5mm}
\caption{Comparison with other published models on the HKUST.}
\vspace{1.5mm}
\begin{tabular}{c|c}
\hline
Model &  CER \\
\hline \hline
Joint CTC-attention model / ESPNet (speed perturb) \cite{watanabe2018espnet} & 27.4  \\
Extended-RNA (speed perturb) \cite{dong2018extending} & 26.8 \\
Transformer (speed perturb) \cite{zhou2018comparison} & 26.6 \\
TDNN-hybrid, lattice-free MMI (speed perturb) \cite{povey2016purely}  & \textbf{23.7}  \\
\hline \hline
SAA model (speed perturb) & 24.1 \\

\hline
\end{tabular}
\label{tab:HkustResults}
\vspace{-0.5mm}
\end{table}

Not only that we find the SAA model obtains a 8.4\%-10.2\% CER reduction than its RNN-based baseline on the CasiaMTS dataset (in table 6), further validating the effectiveness of the SAA.

\begin{table}[!ht]
\centering
\vspace{-2.5mm}
\caption{Comparison on the CasiaMTS dataset.}
\vspace{1.5mm}
\begin{tabular}{l|c|c|c|c}
\hline
Model & Test1 & Test2 & Test3 & Test4 \\
\hline \hline
Extended-RNA \cite{dong2018extending} & 21.20 & 16.63 & 18.10 & 28.81 \\
\hline \hline
SAA model & \textbf{19.13} & \textbf{14.93} & \textbf{16.51} & \textbf{26.38} \\
\hline
\end{tabular}
\label{tab:CasiaMTSResults}
\vspace{-0.5mm}
\end{table}

\section{Conclusion}
\label{sec:print}
%In this work, we present a RNN-free end-to-end model, self-attention aligner (SAA), which replaces RNNs with the SANs in a simplified RNA framework. A joint training mechanism with a pre-trained SAN-LM, and a chunk-hopping mechanism enabling online recognition are also presented for further improving the SAA model. Compared with a RNN-based baseline on Mandarin ASR datasets, we find the SAA (1) obtains a 8.4\%-10.2\% CER reduction; (2) achieves faster calculation speed during training and inference; (3) support latency-control recognition with little performance degradation. These improvements demonstrate the effectiveness of replacing RNNs by the SANs in ASR field. In the future, we will further validate the performance of our SAA model on English ASR task.

In this work, we conduct exploration on the replacement of RNNs by the self-attention networks (SANs) in a simplified RNA framework. We find the SANs could (1) effectively represent speech utterance with temporal down-sampling in the encoder; (2) be compatible with CTC-like alignment mechanism in the decoder. We term the constructed RNN-free model as self-attention aligner (SAA). Compared with a RNN-based baseline on two Mandarin conversation telephone ASR datasets, the SAA model (1) obtains a 8.4\%-10.2\% CER reduction; (2) achieves faster calculation speed during training and inference; (3) supports latency-control recognition with little performance degradation. These advantages demonstrate the effectiveness of replacing RNNs by the SANs in ASR field.

\vfill\pagebreak

\bibliographystyle{IEEEbib}
\bibliography{strings,refs}

\end{document}